# RELATIVE ENTROPY, PROBABILISTIC INFERENCE, AND AI

John E. Shore

Computer Science and Systems Branch
Code 7591
Information Technology Division
Naval Research Laboratory
Washington, D. C. 20375

## I. INTRODUCTION

For probability distributions **q** and **p**, the relative entropy

$$H(\mathbf{q},\mathbf{p}) = \sum_{i=1}^{n} q_i \log \frac{q_i}{p_i} \tag{1}$$

is an information-theoretic measure of the dissimilarity between $\mathbf{q} = q_1, \cdots, q_n$ and $\mathbf{p} = p_1, \cdots, p_n$ ($H$ is also called cross-entropy, discrimination information, directed divergence, I-divergence, K-L number, among other terms). Various properties of relative entropy have led to its widespread use in information theory. These properties suggest that relative entropy has a role to play in systems that attempt to perform inference in terms of probability distributions. In this paper, I will review some basic properties of relative entropy as well as its role in probabilistic inference. I will also mention briefly a few existing and potential applications of relative entropy to so-called artificial intelligence (AI).

## II. INFORMATION, ENTROPY, AND RELATIVE-ENTROPY

### A. Information and Entropy

Suppose that some proposition $x_i$ has probability $p(x_i)$ of being true. Then the amount of information that we would obtain if we learn that $x_i$ is true is

$$I(x_i) = -\log p(x_i). \tag{2}$$

$I(x_i)$ measures the uncertainty of $x_i$. Except for the base of the logarithm, this common definition arises essentially from the requirement that the information implicit in two independent propositions be additive. In particular, if $x_i$ and $x_j$, $i \neq j$, are independent propositions with joint probability $p(x_i \wedge x_j) = p(x_i)p(x_j)$, then the requirements

$$I(x_i \wedge x_j) = I(x_i) + I(x_j) \tag{3}$$

and

$$I(x_i) \geq 0 \tag{4}$$

are sufficient to establish (2) except for the arbitrary base of the logarithm, which determines the units. If the base 2 is used, the resulting units are bits. Equations (3) and (4) are said to provide an *axiomatic characterization* of $I$. If one believes that (3) and (4) are reasonable requirements for an information measure, then one should use (2) because any other measure will violate either (3) or (4).

Let $\{x_i : i = 1, \cdots, n\}$ be a set of mutually exclusive propositions. If we are to learn that one of them is true, then the expected amount of information gained is

$$I(X) = -\sum_{i=1}^{n} p(x_i)I(x_i) = -\sum_{i=1}^{n} p(x_i)\log p(x_i). \tag{5}$$

This quantity, commonly written as $H(\mathbf{p})$, is the *entropy* of the distribution $\mathbf{p}$, and it quantifies the average uncertainty implicit in $\mathbf{p}$. The entropy of $\mathbf{p}$ takes on its maximum value when all of the probabilities are equal, $p_i = p(x_i) = 1/n$, in which case

$$I(X) = \log n \ . \tag{6}$$

Intuitively, this corresponds to the maximum uncertainty being expressed by a uniform distribution. Like the information $I(x_i)$, the entropy is additive. That is, if $\{y_j : j=1, \cdots, m\}$ is a set of mutually exclusive propositions all independent of $\{x_i\}$, then the entropy of the joint distribution $p(x_i, y_j) = p(x_i)p(y_j)$ is additive:

$$I(XY) = I(X) + I(Y) \ . \tag{7}$$

Like the information $I(x_i)$, the entropy $I(X)$ can be characterized axiomatically by means of properties such as (6) and (7) [1].

### B. Information Gain and Relative Entropy

Suppose one acquires evidence $y_j$ about the proposition $x_i$, and as a result the probability of $x_i$ changes from $p(x_i)$ to $p(x_i \mid y_j)$. The information gained is given by the change in uncertainty,

$$I(x_i \ ; y_j) = I(x_i) - I(x_i \mid y_j) = \log \frac{p(x_i \mid y_j)}{p(x_i)} \ . \tag{8}$$

The expected amount of information that $y_j$ provides about $X$ is then

$$I(X \ ; y_j) = \sum_{i=1}^{n} p(x_i \mid y_j) I(x_i \ ; y_j) = \sum_{i=1}^{n} p(x_i \mid y_j) \log \frac{p(x_i \mid y_j)}{p(x_i)} \ . \tag{9}$$

This is the *relative entropy* between the posterior distribution $p(x_i \mid y_j)$ and the prior distribution $p(x_i)$ (see (1)). If we take an additional expectation over $y_j$, the result is the *mutual information*

$$I(X \ ; Y) = \sum_{i=1}^{m} p(y_j) I(X \ ; y_j) = \sum_{ij} p(x_i, y_j) \log \frac{p(x_i, y_j)}{p(x_i)p(y_j)} \ . \tag{10}$$

Note that the mutual information is itself a relative entropy between the joint distribution and the product of the marginals.

The derivation of (9) makes clear the interpretation of relative-entropy as a measure of information dissimilarity, an interpretation that is supported by various properties. For example,

$$H(\mathbf{q},\mathbf{p}) \geq 0 \ , \tag{11}$$

with equality holding if and only if $\mathbf{p} = \mathbf{q}$, a property known as *positivity*. Another property is analogous to (7). Let $p_i = p(x_i)$ and $p'_i = p'(x_i)$ be two distributions on the propositions $x_i$, and let $q_i = p(y_i)$ and $q'_i = p'(y_i)$ be two distributions on the propositions $y_i$. Then

$$H(\mathbf{q}\mathbf{q}',\mathbf{p}\mathbf{p}') = H(\mathbf{q},\mathbf{p}) + H(\mathbf{q}',\mathbf{p}') \tag{12}$$

holds, a property known as *additivity*.

For continuous probability density functions $q(x)$ and $p(x)$, the discrete form of relative entropy (1) becomes

$$H(q,p) = \int q(x) \log \frac{q(x)}{p(x)} \ dx \tag{13}$$

which satisfies the obvious generalizations of (11) and (12). Moveover, the continuous form of relative entropy is invariant under coordinate transformations, another desirable property for an information measure. In contrast, the continuous form of entropy is not coordinate transformation invariant. Like entropy, relative entropy can be characterized axiomatically in terms of properties like (11)-(13), both in the discrete case [2] and in the continuous case [3].

## III. THE PRINCIPLE OF MINIMUM RELATIVE ENTROPY

Let $\mathbf{X}$ be the set of all possible probability distributions on the propositions $\{x_i\}$, and let $\mathbf{q}^\dagger \in \mathbf{X}$ be the actual, unknown probabilities of the $\{x_i\}$. Suppose that $\mathbf{p} \in \mathbf{X}$ is an *initial estimate* of $\mathbf{q}^\dagger$, and suppose that new information about $\mathbf{q}^\dagger$ is obtained in the form of *constraints* that restrict $\mathbf{q}^\dagger$ to some convex subset $\mathbf{I} \subseteq \mathbf{X}$. The set $\mathbf{I}$ is known as the *constraint set*. One example of such constraints is a set of known expected values

$$\sum_{i=1}^{n} q_i^\dagger f_{ki} = \overline{f}_k \quad (k = 1,...,M), \tag{14}$$

or bounds on such expected values. The principle of minimum relative entropy (MRE) states that a *final estimate* of $\mathbf{q}^\dagger$ is obtained from the information $\mathbf{q}^\dagger \in \mathbf{I}$ and the initial estimate $\mathbf{p}$ by finding the distribution $\mathbf{q}$ that minimizes the relative entropy

$$H(\mathbf{q},\mathbf{p}) = \min_{\mathbf{q}' \in \mathbf{I}} H(\mathbf{q}',\mathbf{p}) . \tag{15}$$

Intuitively, this corresponds to choosing the distribution that satisfies the new information $\mathbf{q}^\dagger \in \mathbf{I}$ but is otherwise as close as possible to the initial estimate. For constraints of the form (14), the final estimate $\mathbf{q}$ has the form ( [4, 5, 6] )

$$q_i = p_i \exp\left[-\lambda - \sum_{k=1}^{M} \beta_k f_{ki}\right], \tag{16}$$

where the $\beta_k$ and $\lambda$ are Lagrangian multipliers determined by the constraints (14) and $\sum_i q_i^\dagger = 1$. Conditions for the existence of solutions are discussed by Csiszár [6].

MRE is related to the well-known principle of maximum entropy (ME) [7, 8, 9]. In ME, an estimate of $\mathbf{q}^\dagger$ is obtained by maximizing entropy over the constraint set. No initial estimate is present, and ME corresponds to choosing the estimate that satisfies the constraints but otherwise expresses maximum uncertainty about $\mathbf{q}^\dagger$. MRE can be viewed as a generalization of ME — MRE reduces to ME when the initial estimate is a uniform distribution. The ME solution is given by (17) with the $p_i$ deleted.

Why minimize relative entropy rather than some other function? One answer arises from relative entropy's properties as an information measure. For example, let the symbol $I$ stand for the information $\mathbf{q}^\dagger \in \mathbf{I}$, and let the operator $\circ$ express (15) in the notation

$$\mathbf{q} = \mathbf{p} \circ I . \tag{17}$$

For the case of equality constraints (14), it can then be shown that

$$H(\mathbf{q}^\dagger,\mathbf{p}) = H(\mathbf{q}^\dagger,\mathbf{p} \circ I) + H(\mathbf{p} \circ I,\mathbf{p}) , \tag{18}$$

and it follows that $H(\mathbf{q},\mathbf{p}) = H(\mathbf{p} \circ I,\mathbf{p})$ is the amount of information provided by $I$ that is not inherent in $\mathbf{p}$ [5]. Other answers can be based on frequency of occurrance arguments [10], and on arguments based on consistency with Bayes law [11].

A more fundamental approach is to interpret (17) as a general method of logical inference, to state properties that we require of any consistent method of inference, and then to study their consequences [12, 13]. This amounts to an axiomatic characterization of the operator $\circ$. Here is an informal description of one set of axioms (see [12] for precise statements):

    I.    *Uniqueness:* The result should be unique.

    II.    *Invariance:* The choice of coordinate system should not matter.

    III.    *System Independence:* It should not matter whether one accounts for independent information about independent systems separately in terms of different densities or together

in terms of a joint density.

IV. *Subset Independence:* It should not matter whether one treats an independent subset of system states in terms of a separate conditional density or in terms of the full system density.

It is remarkable that the formal statement of these consistency axioms is sufficient to restrict the form of $H$ in (18) to forms that always yield the same result as the relative entropy (1) [12].

As a general method of statistical inference, MRE minimization was first introduced by Kullback [4]. MRE and ME have been applied in many fields (for a list of references, see [12]). Recent successful applications include spectrum analysis [14, 15, 16], pattern classification [17], speech coding [18], and speech recognition [19].

## IV. RELATIVE ENTROPY AND AI

Although AI emphasizes symbolic information processing, numerical information processing will always have an important role, particular where uncertain information is involved. In this context, there seem to be two areas where relative entropy should have a role to play. The first arises from relative entropy's properties as an information measure — it should be useful as a means of quantifying information gains and losses within probabilistic inference procedures. The second arises from MRE's properties as a uniquely consistent inference procedure, which suggests that MRE can be used directly for inference in AI applications. Indeed, so-called Boltzmann machines and simulated annealing methods [20] can be formulated as applications of MRE, and MRE has been proposed as a means of providing updating in expert systems [21].

A related area of possible application is the problem of searching large state spaces, where one must exploit partial information about some optimal path or node in order to find it or some slightly-suboptimal substitute. One possibility is to use information measures to guide a heuristic search. Another is to use them as a basis for some non-heuristic, information-theoretic search procedure. A hint of this can be found in Karp and Pearl's pruning algorithm for finding a near-optimal path in a binary tree with random costs [22]. In their analysis, the cost of any branch is 1 or 0 with probabilities $p$ and $1-p$. For $p > 1/2$, their algorithm involves a special constant $\alpha^*$ defined by $G(\alpha^*,p) = 1/2$, where

$$G(\alpha,p) = \left(\frac{p}{\alpha}\right)^\alpha \left(\frac{1-p}{1-\alpha}\right)^{1-\alpha}.$$

In fact, $G(\alpha,p) = e^{H(\alpha,p)}$, and $H(\alpha^*,p) = \log 2$; i.e., one bit of information!

## ACKNOWLEDGMENT

Rod Johnson pointed out that $\log G$ is a relative entropy.